\documentclass[conference]{IEEEtran}
\IEEEoverridecommandlockouts

\usepackage{amsmath,amssymb,amsfonts}
\usepackage{graphicx}
\usepackage{xcolor}
\usepackage[caption=false,font=footnotesize]{subfig}
\usepackage{diagbox}
\usepackage[linesnumbered,ruled,vlined,commentsnumbered]{algorithm2e}
\usepackage{algpseudocode}
\usepackage{framed}
\usepackage{multicol}
\usepackage{multirow}
\usepackage[normalem]{ulem}
\usepackage{enumerate}
\usepackage{csquotes}
\usepackage{wasysym}
\usepackage[sorting=none]{biblatex}
\usepackage[colorlinks,citecolor=red,urlcolor=blue,bookmarks=false,hypertexnames=true]{hyperref}

\newcommand{\rn}{\mathbb{R}^n}
\newcommand{\rM}{\mathbb{R}^m}
\newcommand{\rN}{\mathbb{R}^N}
\newcommand{\Rmn}{\mathbb{R}^{m\times n}}
\newcommand{\RNn}{\mathbb{R}^{N\times n}}
\newcommand{\st}{\mathcal{S}}\newcommand{\sgn}{\mathrm{sign}}

\def\BibTeX{{\rm B\kern-.05em{\sc i\kern-.025em b}\kern-.08em
    T\kern-.1667em\lower.7ex\hbox{E}\kern-.125emX}}
\begin{document}

\title{How to warm-start your unfolding network}

\author{
    \IEEEauthorblockN{Vicky Kouni$^{\dagger,\ast}$}
    \IEEEauthorblockA{$^{\dagger}$ \textit{Mathematical Institute, University of Oxford, kouni@maths.ox.ac.uk}}
    \IEEEauthorblockA{$^{\ast}$ \textit{Isaac Newton Institute for Mathematical Studies, University of Cambridge, vk428@cam.ac.uk}} 
    \thanks{The author would like to thank the Isaac Newton Inst. for Math. Sci. for supporting her during her INI Postdoctoral Research Fellowship in the Mathematical Sciences, especially during the programme ``Representing, calibrating \& leveraging prediction uncertainty from statistics to machine learning''. This work was funded by the EPSRC (Grant Number EP/V521929/1).}
}

\maketitle

\begin{abstract}
We present a new ensemble framework for boosting the performance of overparameterized unfolding networks solving the compressed sensing problem. We combine a state-of-the-art overparameterized unfolding network with a continuation technique, to warm-start a crucial quantity of the said network's architecture; we coin the resulting continued network C-DEC. Moreover, for training and evaluating C-DEC, we incorporate the log-cosh loss function, which enjoys both linear and quadratic behavior. Finally, we numerically assess C-DEC's performance on real-world images. Results showcase that the combination of continuation with the overparameterized unfolded architecture, trained and evaluated with the chosen loss function, yields smoother loss landscapes and improved reconstruction and generalization performance of C-DEC, consistently for all datasets.
\end{abstract}


\section{Introduction}
\noindent Deep unfolding networks (DUNs) \cite{ampnet}, \cite{ista-net}, \cite{admm-csnet} burst into the deep learning scene \cite{modeldl}, by merging merits of model-based \cite{sasida} and data-driven approaches \cite{asconvsr}. A DUN is formed by casting the iterations of a model-based algorithm as layers of a deep neural network (DNN), solving an inverse problem \cite{pansharpening}, \cite{dunreverb}. A particular example is the compressed sensing (CS) problem \cite{cs}, \cite{star}, which pertains to reconstructing data $x\in\rn$ from incomplete, noisy measurements $y=Ax+e\in\rM$, $m<n$. Given $y$, a DUN implements a \textit{decoder} $h:\rM\mapsto\rn$, so that $h(y)=\hat{x}\approx x$ \cite{deconet}, \cite{admmdad}. As such, the CS reconstruction problem resembles a regression one \cite{ann}.\\
DUNs enjoy a set of advantages compared to standard DNNs \cite{dr2}, \cite{dendnn} for CS: they are interpretable \cite{interp} and incorporate prior knowledge of data structure \cite{amp}, while they significantly reduce the time complexity and increase the reconstruction quality 
\cite{dynamicdun}. Still, boosting the performance of DUNs is an ongoing research direction. To that end, DUNs are infused with elements from deep-learning architectures. For instance, \cite{lstmdun}, \cite{ufc}, and \cite{gandun} explore the combination of DUNs with LSTM-type gates units, attention mechanisms, and generative data priors, respectively. Nevertheless, the deployment of model-based tools is quite unexplored, although its relevance seems natural, by definition of DUNs.\\
In fact, there are optimization techniques for boosting the performance of model-based iterative algorithms, with a prime example being continuation \cite{fpccs}, \cite{fpcmin}, \cite{nesta}, \cite{fastl1}, also known as ``warm-starting'' \cite{warmstart}, \cite{warmipm}. Coarsely speaking, continuation hinges upon making better and better guesses on the values of parameters/variables used in a CS reconstruction algorithm, in order to improve the algorithm's performance. Since continuation appeals to model-based methods, it appears intuitive to also apply it in the unfolding regime.\\
Our work is inspired by \cite{tfocs}, \cite{demun}, \cite{reddun}. In \cite{tfocs}, continuation is utilized to warm-start the proposed model-based CS reconstruction algorithm. Moreover, \cite{demun} explores how model-based choices, e.g., different matrices $A$, affect the behavior of DUNs, while \cite{reddun} introduces a new loss function for training and evaluating the proposed DUN. Similarly, we will leverage model-based techniques to increase efficiency of DUNs. We differentiate our approach by exploiting continuation to warm-start a key quantity of a state-of-the-art overparameterized DUN, namely, DECONET \cite{deconet}, which solves the CS problem. To our knowledge, warm-starting DUNs for improving their performance has not yet been explored. We coin this new framework of continuation applied on DUN as \textit{continued decoder} (C-DEC). In contrast to standard metrics for assessing DUNs' performance, e.g., the mean-squared loss \cite{admmdad}, \cite{unfoldrnn}, \cite{compound}, we train and evaluate C-DEC using the log-cosh loss function \cite{logcosh}, to increase the reconstruction and generalization performance of the proposed DUN. Log-cosh is a well-known metric for regression problems \cite{statlogcosh} -- though never used before in the context of DUNs -- and combines merits of both mean-absolute error and mean-squared error (cf. Sec.~\ref{exp}). Finally, we experiment with C-DEC on real-world images. Results indicate that continuation and deployment of log-cosh as a metric for testing the proposed DUN, improve reconstruction and generalization performance of C-DEC -- and especially in the case of loss landscapes \cite{lossvis}, where continuation and overparameterization induce a ``smoothing'' effect.\\
Our contributions read as follows: a) we propose a new ensemble framework, which comprises of warm-starting a state-of-the-art overparameterized DUN for CS, resulting in a continued DUN dubbed C-DEC, whose performance is measured in terms of the log-cosh loss; the latter has not been used before in the context of DUNs b) we provide empirical evidence confirming the beneficial role of continuation (and overparameterization) in C-DEC's reconstruction and generalization performance, when the log-cosh loss is employed. We also highlight that continuation and overparameterization smooth out C-DEC's loss landscapes; to our knowledge, we are the first to visualize the loss landscapes of DUNs.\\
\textbf{Notation}: For $x\in\mathbb{R},\,\tau>0$, the soft-thresholding operator $\st_\tau:\mathbb{R}\mapsto\mathbb{R}$ and the truncation operator $\mathcal{T}_\tau:\mathbb{R}\mapsto\mathbb{R}$ are defined in closed form as $\st_{\tau}(x)=\sgn(x)\max(0,|x|-\tau)$ and $\mathcal{T}_{\tau}(x)=\sgn(x)\min\{|x|,\tau\}$, respectively. For $x\in\rn$, both $\st_{\tau}(\cdot)$ and $\mathcal{T}_\tau(\cdot)$ act component-wise. For $x\in\mathbb{R}$, the hyperbolic cosine function is given as $\cosh{(x)}=(e^x+e^{-x})/2$.

\begin{algorithm}[ht]
\small
\SetAlgoLined
\SetAlgoNoEnd
\SetKwData{Left}{left}\SetKwData{This}{this}\SetKwData{Up}{up}\SetKwFunction{Union}{Union}\SetKwFunction{FindCompress}{FindCompress}\SetKwInOut{Input}{Input}\SetKwInOut{Output}{Output}
\Input{$y\in\rM$}
\Output{solution $\hat{x}\in\rn$ of \eqref{l1conic}}
Initialize $x_0\in\rn$, $z_0^1\in\rN$, $z_0^2\in\rM$\;
$\theta_0\leftarrow1$, $u_0^1\leftarrow z_0^1,\,u_0^2\leftarrow z_0^2$\;
\For{$k=0,1,\dots$}{
$x_k\leftarrow x_0+\mu^{-1}((1-\theta_k)W^Tu_k^1+\theta_kW^Tz_k^1-(1-\theta_k)A^Tu_k^2-\theta_kA^Tz_k^2)$\;
$z_{k+1}^1\leftarrow\mathcal{T}_{\theta_k^{-1}t_k^1}((1-\theta_k)u_k^1+\theta_kz_k^1-\theta_k^{-1}t_k^1Wx_k)$\;
$z_{k+1}^2\leftarrow\st_{\theta_k^{-1}t_k^2\varepsilon}((1-\theta_k)u_k^2+\theta_kz_k^2-\theta_k^{-1}t_k^2(y-Ax_k))$\;
$u_{k+1}^1\leftarrow(1-\theta_k)u_k^1+\theta_kz_{k+1}^1$\;
$u_{k+1}^2\leftarrow(1-\theta_k)u_k^2+\theta_kz_{k+1}^2$\;
$\theta_{k+1}\leftarrow2/(1+(1+4/(\theta_k)^2)^{1/2})$\;
}
\caption{\cite[Listing 6]{tfocs}}
\label{l1analysis}
\end{algorithm}
\vspace{-0.2cm}
\section{Main results}\label{main}
\noindent\textbf{Model-based CS}: CS pertains to reconstructing $x\in\rn$ from $y=Ax+e\in\mathbb{R}^m$, $m<n$, with $A\in\Rmn$ and $e\in\mathbb{R}^m$, $\|e\|_2\leq\varepsilon$. The data $x$ is assumed to be sparse after the application of a \textit{known} transform $W$, to ensure exact/approximate reconstruction; in this paper, we consider the case of a \textit{redundant} $W\in\RNn$. In
optimization terms, this is translated into minimizing a smooth, sparsity-inducing objective function satisfying a data-fitting term. In other words, a common approach \cite{tfocs} for formulating the CS problem is:
\begin{equation}\label{l1conic}
    \min_{x\in\rn}\|Wx\|_1+\frac{\mu}{2}\|x-x_0\|_2^2\quad\text{s. t.}\quad\|y-Ax\|_2\leq\varepsilon,
\end{equation}
with $x_0\in\rn$ being an initial guess on $x$, and $\mu>0$. A state-of-the-art algorithm solving \eqref{l1conic} is Algorithm~\ref{l1analysis}. The latter takes as input $y$, introduces dual variables $z^1\in\rN$, $z^2\in\mathbb{R}^m$ and step sizes $t^1,\,t^2,\,\theta$, and produces an iterative scheme, which after some iterations outputs $\hat{x}\approx x$.\\
\noindent Due to the appearance of the term $(\mu/2)\|x-x_0\|_2^2$ in \eqref{l1conic}, the performance of the solver depends on the choices of $x_0$ (and $\mu$); to alleviate this effect, one can use continuation. The latter is an algorithmic method deployed to increase the solver's performance and hinges upon iteratively solving \eqref{l1conic} with better and better guesses on $x_0$ (and $\mu$). In other words, continuation solves a sequence of similar but easier problems, using the results of each ``subproblem'' to ``warm start'', i.e., initialize the next one. The combination of Algorithm~\ref{l1analysis} with continuation results in Algorithm~\ref{cont}, presented here slightly differently from the one in \cite{tfocs}, to satisfy our paper's formulation. We focus on updating $x_0$, and keep $\mu$ fixed, since updating the latter is optional \cite{tfocs}. Of particular interest is the fact that continuation is quite efficient when coupled with Algorithm~\ref{l1analysis} and boosts its performance, in terms of reconstruction error \cite{tfocs}.\\
\noindent\textbf{Deep-unfolding CS}: To reformulate Algorithm~\ref{l1analysis} as a DUN, we firstly consider $W$ to be \textit{unknown} and learned by a set $\mathcal{S}=\{(x_i,y_i)\}_{i=1}^s\overset{\text{i.i.d.}}{\sim}\mathcal{D}^s$, with $\mathcal{D}$ being an unknown distribution. Then, by treating the number of iterations $k$ as layers, Algorithm~\ref{l1analysis} is cast as an unfolded architecture called DECONET \cite{deconet}, with outputs of the $k$th layer given by
\begin{align}
    \label{layer1}
    f_1^W(y)&=\sigma^W_1(y)\\
    \label{layerk}
    f_k^W(v)&=D_{k-1}v+\Theta_{k-1}\sigma^W_{k-1}(v),\quad k=2,\dots,L,
\end{align}
with $v\in\mathbb{R}^{(2N+2m)\times1}$ being an auxiliary variable and $L$ the total number of layers. Formally, $y$ is passed through all subsequent layers, but for the sake of readability, we only write the dependence on $v$ here. The matrices $\{D_k,\Theta_k\}_{k\geq1}\in\mathbb{R}^{(2N+2m)\times(2N+2m)}$ and the nonlinear function $\sigma(\cdot)$ depend on the step sizes and the optimization variables of Algorithm~\ref{l1analysis} (see \cite{deconet} for more implementation details). By composing the outputs layer after layer, DECONET implements a decoder $h^W:\mathbb{R}^m\mapsto\rn$ for CS, such that $h^W(y):=h(y)=\hat{x}\approx x$.  The learnable $W$ is shared across all layers, and due to its redundancy, it renders DECONET as an overparameterized DUN. Since both DECONET and Algorithm~\ref{l1analysis} solve \eqref{l1conic}, we can replace the latter with the former inside the continuation loop of Algorithm~\ref{cont}. This tantamounts to a new unfolded architecture, presented in Algorithm~\ref{contdun}, where we combine the continuation loop with DECONET. We coin this hybrid model-based unfolded architecture \textit{continued decoder} (C-DEC) and write $g^W_{x_0}(y)$ for the decoder it implements, to distinguish it from the decoder implemented by DECONET.\\
Given a training sequence of $s$ pair-samples, we aim to measure the difference between $x_i$ and $\hat{x}_i^\star=g^W_{x_0}(y_i)$, $i=1,\dots,s$. To that end, we choose the log-cosh loss function \cite{logcosh}:
\begin{equation}\label{logcosh}
    \mathcal{L}_{\mathrm{train}}=\frac{1}{s}\sum_{i=1}^s\log\cosh(g^W_{x_0}(y_i)-x_i),
\end{equation}
as opposed to DECONET, which is trained and tested with the mean-squared error (MSE) \cite{deconet}. Our choice of log-cosh \cite{statlogcosh} is attributed to its behavior as the MSE close to the origin, and as the mean-absolute error \cite{superres} far from the origin. Hence, log-cosh incorporates properties of both loss functions, which are standard metrics for DUNs' performance. Based on this aspect and the success of continuation when coupled with the iterative solver for CS, we believe that C-DEC will demonstrate an improved performance compared to its vanilla counterpart. In Sec.~\ref{exp}, we provide empirical evidence, confirming the superiority of the proposed ensemble framework.
\begin{algorithm}[ht]
\small
\SetAlgoLined
\SetAlgoNoEnd
\SetKwData{Left}{left}\SetKwData{This}{this}\SetKwData{Up}{up}\SetKwFunction{Union}{Union}\SetKwFunction{FindCompress}{FindCompress}\SetKwInOut{Input}{Input}\SetKwInOut{Output}{Output}
\Input{$y\in\rM$}
\Output{Warm-started solution $\hat{x}^\star\in\rn$ of \eqref{l1conic}}
Initialize $x_0^0\in\rn$, $\mu_0>0$\;
\For{$j=0,1,\dots$}{
$\hat{x}^{j+1}\leftarrow$Algorithm~\ref{l1analysis}($y$)\;
$x_0^{j+1}\leftarrow\hat{x}^j+\frac{j}{j+3}(\hat{x}^{j+1}-\hat{x}^j)$\;
(optional) linear increase or decrease of $\mu_j$\;}
\caption{Alg.~\ref{l1analysis} with continuation \cite[Listing 11]{tfocs}}
\label{cont}
\end{algorithm}
\begin{algorithm}[b]
\small
\SetAlgoLined
\SetAlgoNoEnd
\SetKwData{Left}{left}\SetKwData{This}{this}\SetKwData{Up}{up}\SetKwFunction{Union}{Union}\SetKwFunction{FindCompress}{FindCompress}\SetKwInOut{Input}{Input}\SetKwInOut{Output}{Output}
\Input{$y\in\rM$}
\Output{Warm-started solution $\hat{x}^\star\in\rn$ of \eqref{l1conic}}
Initialize $x_0^0\in\rn$\;
\For{$j=0,1,\dots$}{
$\hat{x}^{j+1}\leftarrow h(y)$\;
$x_0^{j+1}\leftarrow\hat{x}^j+\frac{j}{j+3}(\hat{x}^{j+1}-\hat{x}^j)$\;
}
\caption{C-DEC}
\label{contdun}
\end{algorithm}
\begin{figure*}[ht]
\centering
\includegraphics[width=0.75\linewidth]{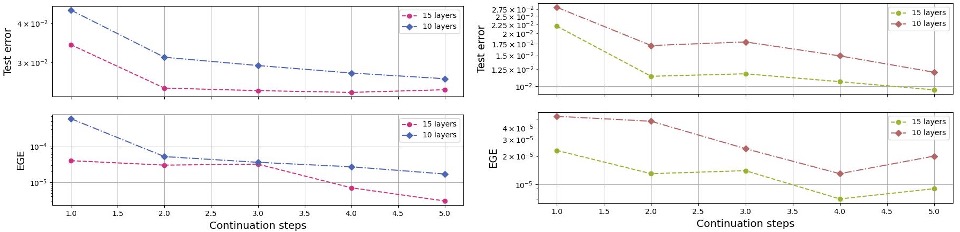}
\caption{Performance plots of C-DEC, with $W\in\mathbb{R}^{10n\times n}$, on MNIST (left) and CIFAR10 (right) datasets.}
\label{perfomplots}
\end{figure*}
\vspace{-0.2cm}
\section{Experiments}
\label{exp}
\noindent\textbf{Settings}: We train and test C-DEC on MNIST (60000 training and 10000 test $28\times28$ image examples) and CIFAR10 (50000 training and 10000 test $32\times32$ coloured image examples). We transform the CIFAR10 images into grayscale ones. For both datasets, we consider the vectorized form of the images. We fix $m=0.25\cdot n$, use a Gaussian $\tilde{A}:=A/\sqrt{m}\in\mathbb{R}^{m\times n}$, and add zero-mean Gaussian noise $e$ with standard deviation $\mathrm{std}=10^{-4}$ to the measurements $y$, so that $y=\tilde{A}x+e$. We set $\varepsilon=\|y-\tilde{A}x\|_2$ and $x_0=A^Ty$, which are standard algorithmic setups for Algorithm~\ref{l1analysis}. We investigate C-DEC with varying number of layers and continuation steps. For the learnable sparsifying transform $W\in\RNn$ we consider two instances, with $N=10\cdot n$ and $N=50\cdot n$, both initialized with a Beta distribution \cite{dyingrelu}. The rest of C-DEC's hyperparameters, e.g., $\mu$, $t^1,t^2,\theta$, are inherited by DECONET and thus set according to \cite{deconet}. All models are implemented in PyTorch \cite{pytorch} and trained using \emph{Adam} algorithm \cite{adam}, with batch size $128$ and learning rates $\eta=10^{-3}$ for MNIST and $\eta=10^{-4}$ for CIFAR10. For our experiments, we report the test log-cosh loss $\mathcal{L}_{\mathrm{test}}$ -- which is essentially the log-cosh loss evaluated on a set of $d$ test data not used during training -- and the \emph{empirical generalization error} (EGE) $\mathcal{L}_{\mathrm{gen}}=|\mathcal{L}_{\mathrm{test}}-\mathcal{L}_{\mathrm{train}}|$, with $\mathcal{L}_{\mathrm{train}}$ as in \eqref{logcosh}. We train C-DEC, on all datasets, employing an early stopping technique \cite{earlystop} with respect to $\mathcal{L}_{\mathrm{gen}}$. We repeat all the experiments at least 10 times and average the results over the runs. Since our main objective is to showcase how the model-based continuation ripples out to the behavior of the unfolded network, we only employ DECONET as a baseline and leave other DUNs as future work, especially since \cite{deconet} has demonstrated that DECONET outperforms two other state-of-the-art DUNs. For the loss landscapes, we follow the standard framework of \cite{lossvis} and perturb the learned $W$ in two random directions, with scalars $Alpha$ and $Beta$ representing the extent of the perturbation in each direction.
\begin{figure*}
    \centering
    \subfloat[Top: $W\in\mathbb{R}^{10n\times n}$. Bottom: $W\in\mathbb{R}^{50n\times n}$.]{
        \includegraphics[width=0.7\textwidth]{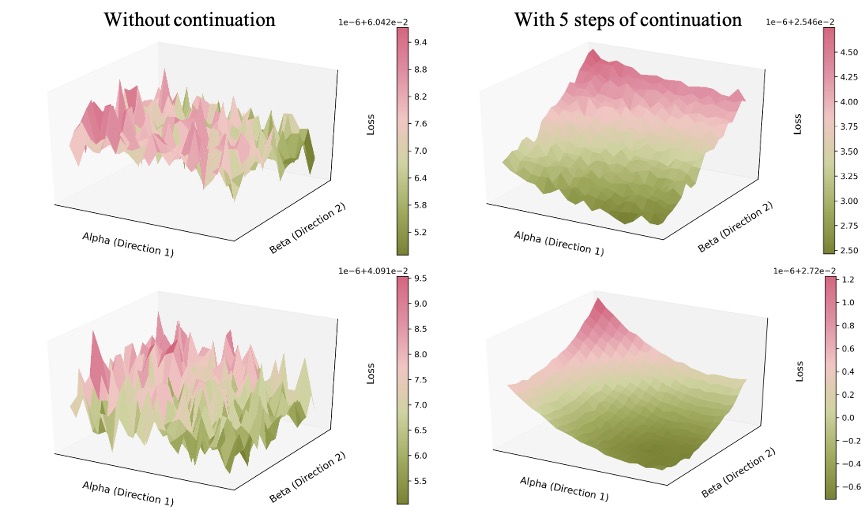}
        \label{mnistlosses}
    }
    \hfill
    \subfloat[Top: $W\in\mathbb{R}^{10n\times n}$. Bottom: $W\in\mathbb{R}^{50n\times n}$.]{
        \includegraphics[width=0.7\textwidth]{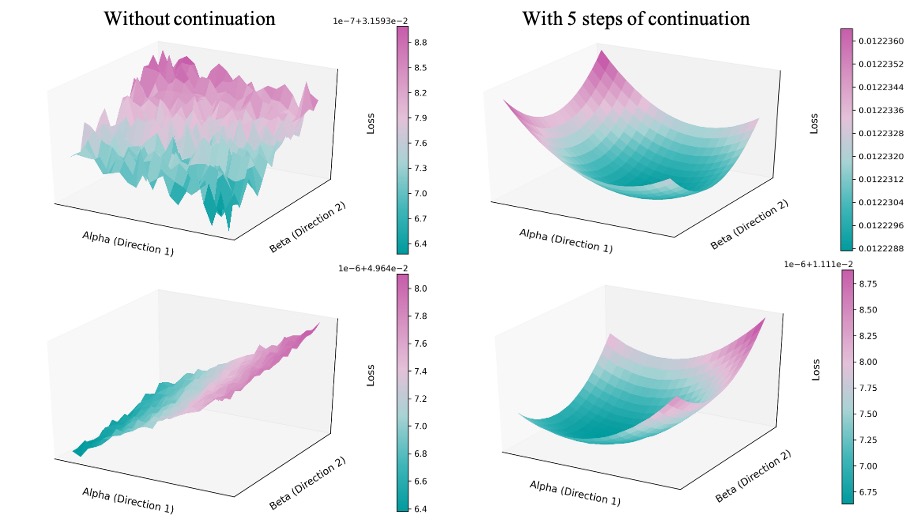}
        \label{cifarlosses}
    }
    \caption{Loss landscapes of 5-layer C-DEC on (a) MNIST and (b) CIFAR10 datasets.}
    \label{lossesplots}
\end{figure*}
\begin{table}[b]
\caption{Performance comparisons between DECONET (measured with the MSE) and C-DEC (measured with the log-cosh loss), on all datasets, with $W\in\mathbb{R}^{50n\times n}$, $L=10$ (top) and $L=50$ (bottom). Bold letters indicate the best performance.}
    \centering
    \scalebox{0.8}{\begin{tabular}{||c|c|c|c|c|c||}
    \hline
     & \multicolumn{2}{|c|}{Test error} & \multicolumn{2}{|c|}{Emp. gen. error}\\
    \hline
    \diagbox{DUN}{Dataset} & MNIST & CIFAR10 & MNIST & CIFAR10\\
    \hline\hline
    C-DEC (proposed) & \textbf{0.029893} & \textbf{0.014344} & \textbf{0.000018} & \textbf{0.000008}\\
    \hline
    DECONET & 0.055963 & 0.038947 & 0.000214 & 0.000139\\
    \hline
    \end{tabular}}
    
    \medskip
    
    \scalebox{0.8}{\begin{tabular}{||c|c|c|c|c|c||}
    \hline
     & \multicolumn{2}{|c|}{Test error} & \multicolumn{2}{|c|}{Emp. gen. error}\\
    \hline
    \diagbox{DUN}{Dataset} & MNIST & CIFAR10 & MNIST & CIFAR10\\
    \hline\hline
    C-DEC (proposed) & \textbf{0.025576} & \textbf{0.013918} & \textbf{0.000086} & \textbf{0.000074}\\
    \hline
    DECONET & 0.061135 & 0.027518 & 0.000117 & 0.000161\\
    \hline
    \end{tabular}}
    \label{losses}
\end{table}

\noindent\textbf{Results \& discussion}: We examine the reconstruction and generalization performance of 10- and 15-layer C-DEC, with $W\in\mathbb{R}^{10n\times n}$ and varying number of continuation steps, on both datasets, and illustrate the findings in Fig.~\ref{perfomplots}. From a model-based viewpoint, the decays in test errors seem reasonable: as the number of continuation steps (and layers) increases, the solver's reconstruction ability is improved \cite{tfocs}. This behavior conforms with our intuition presented in Sec.~\ref{main}. Interestingly, we find that the number of continuation steps is mildly reversely proportional to the generalization error. We surmise that this is due to the nested architecture of C-DEC compared to DECONET, although further mathematical exploration is needed. Additionally, we compare C-DEC and DECONET, both with $W\in\mathbb{R}^{50n\times n}$, for 10 and 50 layers, on all datasets, and present the findings in Table~\ref{losses}. Both the test and generalization errors are always lower for our proposed DUN, consistently for both datasets. Based on the results, we obtain a two-fold insight: a) we confirm our model-based intuition that continuation boosts the solver's performance, even in the unfolding regime b) we justify our choice of the log-cosh loss as an adequate metric for training and evaluating the proposed DUN, compared to MSE, which is deployed for DECONET. Overall, results highlight that continuation is a favorable component of C-DEC's architecture, with the log-cosh loss being a better choice over standard metrics, e.g., the MSE, for measuring DUNs' performance.\\
Finally, we examine how continuation ripples out to C-DEC's loss landscapes. We compare a 5-layer C-DEC without continuation -- which is architecturally equivalent to DECONET -- and a 5-layer C-DEC with 5 steps of continuation. We examine both DUNs with $W\in\mathbb{R}^{10n\times n}$ and $W\in\mathbb{R}^{50n\times n}$, and report the results in Fig.~\ref{lossesplots}. The illustrations demonstrate a very intriguing phenomenon. Firstly, we observe that for both datasets, DECONET (equivalent to C-DEC with 1-step continuation) suffers from a highly nonsmooth loss landscape, with many peaks and valleys. On the other hand, C-DEC with 5 steps of continuation enjoys a much smoother loss landscape, with the effect being highly noticeable for the CIFAR10 images, as depicted in Fig.~\ref{cifarlosses}. Then, for the MNIST dataset illustrated in Fig.~\ref{mnistlosses}, we notice that overparameterization also smooths out the landscape, as $N$ increases from $10n$ to $50n$. Although this effect is negligible for DECONET, it is more noticeable for C-DEC, showcasing the potential of coupling continuation with overparameterization in the context of DUNs. All in all, results indicate that continuation (and overparameterization) has a strong positive effect on the optimization process of DUNs, and could spark an interesting mathematical theory regarding DUNs' generalization.

\section{Conclusion}
In this paper, we applied a continuation technique to warm-start a state-of-the-art overparameterized unfolding network solving the compressed sensing problem. We measured the performance of the resulting continued network with an enhanced loss function, as opposed to standard metrics for unfolding networks. Empirical results highlighted the superiority of the proposed framework, thereby paving the way for creating hybrid unfolded architectures, with tools stemming from model-based methods. In the future, we would like to mathematically explore the properties of the proposed network. This could include a comprehensive optimization-based analysis similar to \cite{lossover}, or the study of continuation's effect on the network's robustness and generalization ability.

\clearpage

\printbibliography

\end{document}